\documentclass{article}

\usepackage{microtype}
\usepackage{graphicx}
\usepackage{subcaption}
\usepackage{booktabs}

\usepackage{hyperref}
\usepackage{forest}

\usepackage[preprint]{icml2026}

\usepackage[
  backend=biber,
  style=authoryear-comp,
  maxcitenames=2,
  maxbibnames=99,
  giveninits=true,
  natbib=true,
  uniquename=false,
  uniquelist=false,
]{biblatex}
\renewbibmacro*{cite:labelyear+extrayear}{%
  \iffieldundef{labelyear}
    {}
    {\printfield{labelyear}%
     \printfield{extrayear}}}
\renewbibmacro*{cite}{%
  \iffieldundef{shorthand}
    {\printnames{labelname}%
     \setunit{\printdelim{nameyeardelim}}%
     \iffieldundef{labelyear}
       {}
       {\printfield{labelyear}%
        \printfield{extrayear}}}
    {\usebibmacro{cite:shorthand}}}
\renewbibmacro*{textcite}{%
  \iffieldundef{shorthand}
    {\printnames{labelname}%
     \setunit{\printdelim{nameyeardelim}}%
     \printtext[parens]{%
       \iffieldundef{labelyear}
         {}
         {\printfield{labelyear}%
          \printfield{extrayear}}}}
    {\usebibmacro{cite:shorthand}}}
\DeclareCiteCommand{\textcite}
  {\boolfalse{cbx:parens}}
  {\usebibmacro{citeindex}%
   \printtext[bibhyperref]{\usebibmacro{textcite}}}
  {\ifbool{cbx:parens}
     {\bibcloseparen\global\boolfalse{cbx:parens}}
     {}%
   \multicitedelim}
  {\usebibmacro{textcite:postnote}}
\DeclareCiteCommand{\parencite}[\mkbibparens]
  {\usebibmacro{prenote}}
  {\usebibmacro{citeindex}%
   \printtext[bibhyperref]{\usebibmacro{cite}}}
  {\multicitedelim}
  {\usebibmacro{postnote}}
\DeclareCiteCommand{\citealp}
  {\usebibmacro{prenote}}
  {\usebibmacro{citeindex}%
   \printtext[bibhyperref]{\usebibmacro{cite}}}
  {\multicitedelim}
  {\usebibmacro{postnote}}
\AtEveryBibitem{\clearfield{month}\clearfield{note}}

\usepackage{amsmath,
amssymb,
amsthm,
mathtools}

\usepackage{tikz-cd}
\usetikzlibrary{positioning,arrows.meta}

\newcommand{\kyle}[1]{}
\newcommand{\nate}[1]{}

\usepackage[capitalize,noabbrev]{cleveref}

\theoremstyle{plain}
\newtheorem{theorem}{Theorem}[section]

\theoremstyle{definition}
\newtheorem{definition}[theorem]{Definition}

\theoremstyle{remark}
\newtheorem{remark}[theorem]{Remark}
\theoremstyle{definition}
\newtheorem{example}[theorem]{Example}

\newcommand\cA{\mathcal{A}}

\newcommand\cF{\mathcal{F}}

\newcommand\cK{\mathcal{K}}

\newcommand\cO{\mathcal{O}}

\newcommand\cQ{\mathcal{Q}}

\newcommand\cV{\mathcal{V}}

\newcommand{\bR}{\mathbb{R}}

\newcommand{\Set}{\textsf{Set}}

\newcommand{\ivltwo}[2]{%
  \tikz[baseline=-0.5ex, x=1cm, y=1cm]{%
    \draw[thick] (0,0)--(1.2,0);
    \foreach \t in {0,#1,#2,1.2}{\draw[thick](\t,-0.07)--(\t,0.07);}
  }%
}
\newcommand{\ivlthree}[3]{%
  \tikz[baseline=-0.5ex, x=1cm, y=1cm]{%
    \draw[thick] (0,0)--(1.2,0);
    \foreach \t in {0,#1,#2,#3,1.2}{\draw[thick](\t,-0.07)--(\t,0.07);}
  }%
}

\usepackage[textsize=tiny]{todonotes}

\icmltitlerunning{Operads for compositional reasoning in LLMs}

\begin{document}

\twocolumn[
  \icmltitle{Operads for compositional reasoning in LLMs}

  \icmlsetsymbol{equal}{*}

  \begin{icmlauthorlist}
    \icmlauthor{Nathaniel Bottman}{incub}
    \icmlauthor{Kyle Richardson}{ai2}
  \end{icmlauthorlist}

  \icmlaffiliation{incub}{Incubilate, Seattle, USA}
  \icmlaffiliation{ai2}{Allen Institute for Artificial Intelligence, Seattle, USA}

  \icmlcorrespondingauthor{Nathaniel Bottman}{nate@incubilate.com}

  \vskip 0.3in
]

\printAffiliationsAndNotice{}

\begin{abstract}
Question decomposition, i.e. breaking a complex query into simpler sub-queries whose answers are composed to produce a final answer, is a widely used strategy for improving LLM reasoning, yet it currently lacks a rigorous mathematical foundation. In this paper, we propose operads, mathematical structures that model many-in, one-out operations and compositions thereof, as a natural framework for describing question decomposition. We define the questions operad $\mathcal{Q}$, in which operations correspond to question templates and composition corresponds to substitution of sub-answers, and show how QA models can be interpreted as algebras over $\mathcal{Q}$. Beyond reframing existing practice, this operadic perspective points toward new methods, in particular a notion of operadic consistency, which measures whether a QA model's answers agree across the partial collapses of a question decomposition tree. Empirical evaluation of operadic consistency is reported in our companion paper (Bottman, Liu, and Richardson, 2026), which finds it strongly correlated with accuracy across twelve LLMs and four multi-hop QA datasets and outperforming standard temperature-based self-consistency baselines. We argue that operads are the natural mathematical home for question decomposition, and that invariants such as operadic consistency open new directions for analyzing and improving the reliability of multi-step reasoning.
\end{abstract}

\section{Question decomposition in practice}

Large Language Models (LLMs) have demonstrated remarkable reasoning capabilities, particularly when prompted to break problems down into intermediate steps.
Such an idea was first popularized in work on chain-of-thought prompting \citep{wei2022chain} and its many subsequent extensions \citep{yao2023tree, khot2022decomposed, wang2022self, yang2024buffer, 
besta2024graph}.
In practice, this process involves transforming a complex query into simpler sub-queries whose answers are sequentially composed to derive a final solution --- for instance, answering \emph{How long was it between when the Titanic hit the iceberg and when it sank completely?}\ by first resolving the timestamps of the collision and the sinking.

Despite its intuitive appeal and widespread adoption, question decomposition currently lacks a rigorous mathematical foundation.
This matters: without a formal account of composition structure, it is difficult to define what it means for a decomposition to be correct or well-formed, to reason about how errors propagate through 
a chain of sub-queries, or to compare decomposition strategies in a principled way --- all independently of any particular model implementation.

In this paper, we propose operads as the right mathematical framework for this problem.
Operads were introduced by mathematicians to study systems built by iterated substitution --- how many-input operations compose coherently into larger operations --- and they provide a compact language for tree-shaped composition.
In our setting, operations correspond to question templates with blanks, and composites correspond to decompositions into sub-queries.
Relatedly, \citet{marcolli2023syntax} have applied operadic and associahedral structures to model syntactic Merge and the syntax-semantics interface in generative linguistics, further supporting the view that operads are a natural framework for compositional structure in language.
Beyond developing the formalism, we point to our companion paper \citep{bottman:operadic_consistency_empirical} for an empirical evaluation of operadic consistency across twelve LLMs and four multi-hop QA datasets.

\section{Operads}
\label{s:operads}

Operads are structures for organizing collections of ``morphisms'', i.e.\ operations that take $k\geq0$ inputs and return one output.
They are closely related to \emph{categories}, which are better-known objects that organize one-in, one-out operations.

In this section, we will introduce the reader to operads, and then explain how question decomposition and chain-of-thought can be described in terms of the \emph{questions operad}.
We assume no prior knowledge of category theory or operads, and prioritize accessibility.
Throughout, the objects we call ``operads'' are, more precisely, ``non-symmetric non-unital operads in $\Set$''.

\subsection{Operads in mathematics}
\label{ss:operads_in_math}

We begin this section with a concise explanation of the notion of an operad.
There are few approachable references; we suggest the curious reader consult Markl--Shnider--Stasheff's {\it Operads in Algebra, Topology and Physics} \citep{markl_stasheff_shnider:operads_in_algebra_topology_and_physics}, and Loday--Vallette's {\it Algebraic Operads} \citep{loday_vallette:algebraic_operads}.

Operads model properties of many-in, one-out morphisms.
One feature that such morphisms have is that we can compose them.
Indeed, consider mappings $f$ and $g$, which take $k$ and $\ell$ inputs, respectively.
For any $1 \leq i \leq k$, we can form the composition
\begin{align}
\label{eq:function_composition}
f\Bigl(\underbrace{-,\ldots,-}_{i-1},g\Bigl(\underbrace{-,\ldots,-}_\ell\Bigr),\underbrace{-,\ldots,-}_{k-i}\Bigr),
\end{align}
i.e.\ the result of feeding the output of $g$ into the $i$-th input slot of $f$.
The result is a mapping that takes $k+\ell-1$ inputs, which we denote $f\circ_ig$.
Iterated compositions are unambiguous --- for instance, when we write the expression
\begin{align}
\label{eq:classical_assoc}
f(\ldots,g(\ldots,h(\ldots),\ldots),\ldots),
\end{align}
the meaning is independent of whether we first compose $f$ with $g$ and then feed in $h$, or first compose $g$ with $h$ and then feed the result into $f$.

These observations lead us to the definition of an operad.

\begin{definition}
\label{def:operad}
A \emph{(non-unital) operad $\cO$} is the following data:
\begin{itemize}
\item A collection $\cO(0), \cO(1), \cO(2), \ldots$ of sets.
(Elements of $\cO(k)$ are thought of as arity-$k$ morphisms, i.e.\ operations with $k$ inputs and one output.)

\item For every $k, \ell \geq 0$ and $1 \leq i \leq k$, a ``composition map'' $\circ_i\colon \cO(k) \times \cO(\ell) \to \cO(k+\ell-1)$.
(When $\cO$ is literally a collection of morphisms, and we are given $f \in \cO(k)$ and $g \in \cO(\ell)$, $f \circ_i g$ is obtained by plugging the output of $g$ into the $i$-th input slot of $f$.
The result is a $(k+\ell-1)$-ary operation.)
\end{itemize}
These data must satisfy the following conditions:
\begin{itemize}
\item
Composition is \emph{sequentially associative}.
That is, for every $k, \ell, m \geq 0$, $f \in \cO(k), g\in\cO(\ell), h\in\cO(m)$, and $1 \leq i \leq k$, $1 \leq j \leq \ell$, the compositions
\begin{align}
f \circ_i (g \circ_j h),
\quad
(f \circ_i g) \circ_{i+j-1} h
\end{align}
agree.
(Compare \eqref{eq:classical_assoc}.)

\item
Composition into disjoint slots commutes (\emph{parallel associativity}).
For $f \in \cO(k)$, $g \in \cO(\ell)$, $h \in \cO(m)$, and $1 \leq i < j \leq k$, the compositions
\begin{align}
(f \circ_i g) \circ_{j+\ell-1} h,
\quad
(f \circ_j h) \circ_i g
\end{align}
agree.
(The shifted index $j+\ell-1$ accounts for $g$'s $\ell$ blanks displacing the $j$-th slot of $f$.)
\null\hfill$\triangle$
\end{itemize}
\end{definition}

In the rest of the subsection, we describe two operads with connections to classical computer science and formal language theory.

\begin{example}
Given an alphabet $\Sigma$, we can form the set $\Sigma^*$ of finite strings over $\Sigma$.
We define the \emph{text processing operad over $\Sigma$}, denoted $\cF_\Sigma$, like so:
\begin{itemize}
\item $\cF_\Sigma(k)$ is the set of all mappings from $(\Sigma^*)^k$ to $\Sigma^*$, i.e.\ all mappings that take $k$ inputs, all lying in $\Sigma^*$, and return one element of $\Sigma^*$.

\item Composition is ordinary composition of mappings, as in \eqref{eq:function_composition}.
\end{itemize}
\noindent
For instance, $\cF_\Sigma(0)$ can be identified with $\Sigma^*$ itself; $\cF_\Sigma(1)$ contains the operator sending a string to its all-caps version; $\cF_\Sigma(2)$ contains the binary concatenation operator $(x,y)\mapsto xy$; and $\cF_\Sigma(3)$ contains ternary operations such as the map sending $(x,y,z)$ to the string obtained by joining $x$, $y$, and $z$ with the separator ``$\mathtt{,}$''.
Let $f$ and $g$ denote the binary and ternary operators just introduced, respectively.
Then $f \circ_1 g$ (i.e., $f$ with $g$ plugged into its first input slot) and $f \circ_2 g$ (i.e., $f$ with $g$ plugged into its second input slot) are distinct elements of $\cF_\Sigma(4)$.
\null\hfill$\triangle$
\label{ex:text_processing}
\end{example}

\paragraph{Colored operads}
\emph{Colored} (or \emph{typed}) operads are like operads, except that each morphism takes inputs of specific \emph{colors}, and returns an output of a specific color (or type).
If $\cO$ is a colored operad, we denote by $\cO(c_1,\ldots,c_k;d)$ the morphisms that take $k$ inputs with colors $c_1, \ldots, c_k$, respectively, and return an output of color $d$.
(A category can be interpreted as a colored operad with only unary operations; for this reason, colored operads are sometimes called \emph{multicategories}.)

\begin{example}
\label{ex:D_G}
Suppose that $G = (V,\Sigma,R,S)$ is a context-free grammar, as in \S2.1 of \citet{sipser}.
We define a \emph{derivation tree with non-terminal leaves} for $G$ to be a rooted planar tree $T$ such that:
\begin{itemize}
\item Every node is labeled by a non-terminal $A \in V$;

\item If an internal node is labeled $A$, and its children are labeled $B_1, \ldots, B_k$, then $A \to B_1,\ldots,B_k$ must be a production rule in $R$.
\end{itemize}
We define $D_G$ to be the colored operad of derivation trees with non-terminal leaves for $G$, with colors $V$.
That is:
\begin{itemize}
\item Given $B_1, \ldots, B_k \in V$ and $A \in V$, $D_G(B_1,\ldots,B_k;A)$ consists of the derivation trees for $G$ whose $k$ leaves have labels $B_1, \ldots, B_k$ and whose root has label $A$.

\item Composition in $D_G$ is given by grafting the root of one derivation tree to a leaf of another --- a common operation in context-free parsing --- subject to the requirement that the labels at the two nodes being identified must match.
\end{itemize}
For instance, suppose that $G$ is a context-free grammar modeling English, with non-terminals \{S, NP, VP, PP, Det, N, V, P\} (standing for ``sentence'', ``noun phrase'', etc.).
Here are two derivation trees, called $T_1$ and $T_2$, which define elements of $D_G(\text{Det},\text{N},\text{VP};\text{S})$ and $D_G(\text{V},\text{Det},\text{N},\text{P},\text{Det},\text{N};\text{VP})$, respectively:
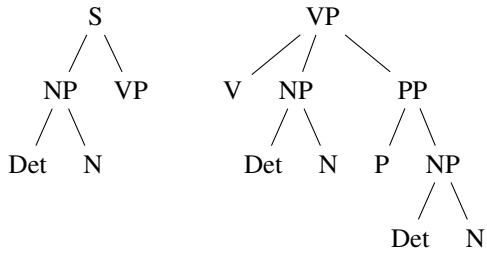
\begin{center}
\begin{forest}
  [, phantom, s sep=0.75cm
    [S
      [NP
        [Det]
        [N]
      ]
      [VP]
    ]
    [VP  
      [V]
      [NP
        [Det]
        [N]
      ]
      [PP
        [P]
        [NP
          [Det]
          [N]
        ]
      ]
    ]
  ]
\end{forest}
\end{center}
We can form the composition $T_1\circ_3T_2$, which grafts the trees at their unique VP node to yield an element of $D_G(\text{Det},\text{N},\text{V},\text{Det},\text{N},\text{P},\text{Det},\text{N};\text{S})$:
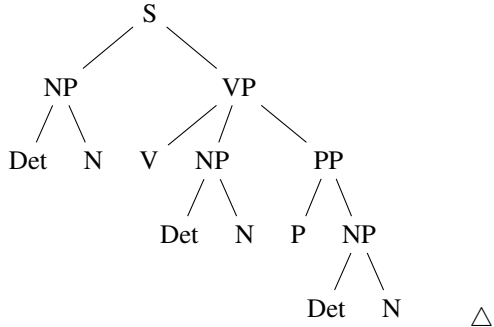
\begin{center}
\begin{forest}
  [S
    [NP
      [Det]
      [N]
    ]
    [VP
      [V]
      [NP
        [Det]
        [N]
      ]
      [PP
        [P]
        [NP
          [Det]
          [N]
        ]
      ]
    ]
  ]
\end{forest}
\qquad$\triangle$
\end{center}
\label{ex:derivation}
\end{example}

\paragraph{Why operads are helpful.}

The appeal of operads is that they allow us to separate the abstract forms of composition from any particular interpretation of those forms.
For example, Examples~\ref{ex:text_processing}--\ref{ex:D_G} talk about very different objects (i.e., $k$-ary string processing functions vs.\ derivation trees in formal grammars), yet operads package both as instances of the same underlying idea: an operation with multiple inputs and one output, together with a notion of plugging one operation into an input slot of another.

Operads were originally introduced by Peter May in algebraic topology \citep{may:recognition}, where one motivating problem was to understand the algebraic structure of loopspaces.
The basic idea of a loop is simple: fix a point $x_0$ in a space $X$, and consider continuous paths $\gamma\colon [0,1] \to X$ that start at $x_0$ (i.e. $\gamma(0)=x_0$), trace out a route through $X$, and return to $x_0$ at the end of their interval (i.e.\ $\gamma(1)=x_0$).
The loopspace $\Omega X$ is the collection of all such loops.

Loopspaces are compositional because one can traverse one loop $\gamma_1$ and then a second $\gamma_2$, resulting in the \emph{concatenation} $\gamma_1*\gamma_2$.
We can depict this concatenation as the unit interval $[0,1]$, with a tick mark indicating the transition from $\gamma_1$ to $\gamma_2$: \hspace{1.0em}\begin{tikzpicture}[baseline=(current bounding box.center), x=0.5cm, y=0.5cm]
  \draw[thick] (0,0)--(2.4,0);
  \foreach \t in {0,1.2,2.4}{\draw[thick](\t,-0.12)--(\t,0.12);}
  \node[below=1pt] at (0.6,-0.12) {\scriptsize$\gamma_1$};
  \node[below=1pt] at (1.8,-0.12) {\scriptsize$\gamma_2$};
\end{tikzpicture}

Concatenation, however, is not associative.
Indeed, $(\gamma_1 * \gamma_2) * \gamma_3$ and $\gamma_1 * (\gamma_2 * \gamma_3)$ traverse the same loops, but do so with different speeds and are therefore distinct as loops.
These concatenations are nevertheless deformable into one another by reparametrizing time, which corresponds to moving the tick marks on the interval.
There are thus two ways to concatenate three loops, related by a single deformation; we organize them into the endpoints of an interval, the \emph{associahedron} $\cK(3)$:

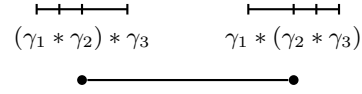
\begin{center}
\begin{tikzpicture}[x=1cm, y=1cm]
  \node[fill=black, circle, inner sep=1.3pt,
    label={[label distance=2mm]above:{\shortstack{
      \ivltwo{0.3}{0.6}\\[1pt]
      \small$(\gamma_1*\gamma_2)*\gamma_3$}}}]
    (l) at (0,0) {};
  \node[fill=black, circle, inner sep=1.3pt,
    label={[label distance=2mm]above:{\shortstack{
      \ivltwo{0.6}{0.9}\\[1pt]
      \small$\gamma_1*(\gamma_2*\gamma_3)$}}}]
    (r) at (2.8,0) {};
  \draw[thick] (l)--(r);
\end{tikzpicture}
\end{center}

For four loops, there are five parenthesizations.
This time, the deformations that relate these concatenations are organized into a pentagon, the associahedron $\cK(4)$:

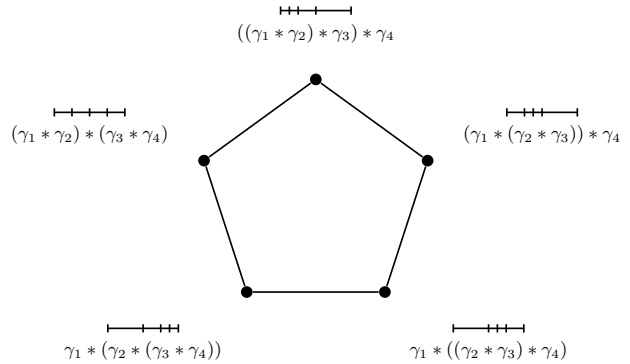
\begin{center}
\resizebox{\columnwidth}{!}{%
\begin{tikzpicture}[x=1cm, y=1cm]
  \def\r{2.0}
  \node[fill=black, circle, inner sep=2pt,
    label={[label distance=4mm]90:{\shortstack{
      \ivlthree{0.15}{0.3}{0.6}\\[1pt]
      \small$((\gamma_1*\gamma_2)*\gamma_3)*\gamma_4$}}}]
    (v0) at (90:\r) {};
  \node[fill=black, circle, inner sep=2pt,
    label={[label distance=4mm]18:{\shortstack{
      \ivlthree{0.3}{0.45}{0.6}\\[1pt]
      \small$(\gamma_1*(\gamma_2*\gamma_3))*\gamma_4$}}}]
    (v1) at (18:\r) {};
  \node[fill=black, circle, inner sep=2pt,
    label={[label distance=4mm]-54:{\shortstack{
      \ivlthree{0.6}{0.75}{0.9}\\[1pt]
      \small$\gamma_1*((\gamma_2*\gamma_3)*\gamma_4)$}}}]
    (v2) at (-54:\r) {};
  \node[fill=black, circle, inner sep=2pt,
    label={[label distance=4mm]234:{\shortstack{
      \ivlthree{0.6}{0.9}{1.05}\\[1pt]
      \small$\gamma_1*(\gamma_2*(\gamma_3*\gamma_4))$}}}]
    (v3) at (234:\r) {};
  \node[fill=black, circle, inner sep=2pt,
    label={[label distance=4mm]162:{\shortstack{
      \ivlthree{0.3}{0.6}{0.9}\\[1pt]
      \small$(\gamma_1*\gamma_2)*(\gamma_3*\gamma_4)$}}}]
    (v4) at (162:\r) {};
  \draw[thick] (v0)--(v1)--(v2)--(v3)--(v4)--(v0);
\end{tikzpicture}}
\end{center}

Unlike the purely combinatorial operad $D_G$, the associahedra are polytopes --- topological spaces whose cells record not merely that parenthesizations are related, but the continuous geometry of how they deform into one another.
One might instead simply identify all parenthesizations as homotopy-equivalent and work with the quotient; but this would discard the finer structure that concatenation naturally carries.

May's insight was that these coherence data are precisely the kind of structure operads encode.
In modern terms, loopspaces are naturally algebras over the associahedra operad, and May's Recognition Principle says, roughly, that under suitable hypotheses this operadic structure characterizes loopspaces. This illustrates the ``operadic mantra'':
\begin{center}
{\it
Objects with extra structure should be understood as algebras over an appropriate operad.}
\end{center}
This example also connects to the grammar and question-decomposition examples below.
The associahedra organize different parenthesizations of a composite expression; derivation trees similarly record different histories for composing constituents into a string; and question decompositions record different histories for composing subquestions into a final query.
In all three cases, the compositional history itself carries information.
Operads provide a language for studying that history, rather than only the final output.

We have two purposes to bring the language of operads to bear in the context of question decomposition and machine learning more generally.
First, we believe that operads are simply the natural formal structure for question decomposition, and therefore that their introduction in this context is innately worthwhile.
Second, reformulating question decompositions and QA models in operadic terms can lead to new results and tools --- for instance, a new notion of ``operadic consistency'' for QA models (see \S\ref{sss:operadic_consistency}).
Even in the well-studied context of context-free grammars, we describe in Remark~\ref{rmk:cfg_ambiguity} how the operadic point of view can point to new tools.

\subsubsection{Algebras over operads}
\label{sss:algebras}

Operads do not necessarily have to literally be collections of operations (for instance, consider $D_G$ from Example~\ref{ex:D_G}).
A realization of an operad in terms of actual operations is called an \emph{algebra}.

\begin{definition}
Suppose that $\cO$ is a colored operad.
An \emph{algebra $\cA$ over $\cO$} is:
\begin{itemize}
\item For every color $c$, a set $\cA(c)$.

\item For every $f \in \cO(c_1,\ldots,c_k;d)$, a map
\begin{align}
\varphi_f\colon \cA(c_1)\times\cdots\times\cA(c_k)\to\cA(d).
\end{align}
\end{itemize}
We require that $\varphi_{f \circ_i g}$ is the same operation as inserting the output of $\varphi_g$ into the $i$-th input of $\varphi_f$; this condition is also referred to as ``associativity''.
\null\hfill$\triangle$
\end{definition}

For instance, $\Sigma^*$ is naturally an algebra over $\cF_\Sigma$: for each $f \in \cF_\Sigma(k)$, the corresponding algebra operation is simply the map $f : (\Sigma^*)^k \to \Sigma^*$.
We describe another example of an algebra, this time over the colored operad $D_G$ from Example~\ref{ex:D_G}, below.

\begin{example}
Given a context-free grammar, we can define an algebra $L_G$ over the operad $D_G$ of derivation trees for $G$, like so:
\begin{itemize}
\item For $A \in V$, $L_G(A)$ is the language generated by $A$, i.e.\ the language of the subgrammar with $A$ as the start symbol.

\item If $T$ is a derivation tree for $G$, with root label $A$ and leaf labels $B_1,\ldots,B_k$, then it defines a map
\begin{align}
\varphi_T\colon
L(B_1)\times\cdots\times L(B_k) \to L(A)
\end{align}
by concatenation.
For instance, the operation associated to the derivation tree described near the end of Example~\ref{ex:D_G} could take as inputs \{``The'', ``dog'', ``saw'', ``the'', ``cat'', ``in'', ``the'', ``park''\} and return the sentence ``The dog saw the cat in the park''.
\null\hfill$\triangle$
\end{itemize}
\end{example}

The content of a context-free grammar can therefore be repackaged as the operad $D_G$ and its algebra $L_G$.
Importantly, $L_G$ is only one possible algebra over $D_G$.
Indeed, besides the string yield algebra, one can define algebras for computing derivation probabilities, best-parse computations, or derivation counts.
This connects to classical work on weighted and semiring parsing \citep{goodman1999semiring, nederhof2003weighted}, where different computations are obtained by varying the value algebra while keeping the underlying derivational machinery fixed.
The operadic viewpoint complements this line of work by treating the grafting structure of derivations as a first-class algebraic object.
This is useful not only for computing values over derivations, but also for defining more complex algebraic constructions that capture properties of the derivational system itself, such as the ambiguity algebra introduced below and, later, our notion of operadic consistency for question decomposition.

\begin{remark}
\label{rmk:cfg_ambiguity}
At least as early as the foundational work of \citet{chomsky_schuetzenberger}, \emph{ambiguity} has been regarded as a central notion in the theory of context-free grammars and languages.
One treatment can be found in \S5.4 of \citet{hopcroft_motwani_ullman:introduction_to_automata}, where the authors define a CFG $G$ to be unambiguous if each string has at most one derivation tree in $G$.
This definition is binary and has no internal algebraic structure.

Our reformulation of a CFG as an operad $D_G$ and an algebra $L_G$ enables us to define a new avatar of ambiguity, called the \emph{ambiguity algebra $\ker\varphi$}.
This is a canonical object that recovers the usual definition of ambiguity as the condition $\ker\varphi \neq 0$, while additionally encoding the compositional structure of ambiguity.

To define the ambiguity algebra, we first linearize $D_G$ and $L_G$ to form $D_G^\bR$ and $L_G^\bR$.
That is, we define $D_G^\bR(B_1,\ldots,B_k;A)$ to consist of formal $\bR$-linear combinations of elements of $D_G(B_1,\ldots,B_k;A)$, and $L_G^\bR(A)$ similarly.
$L_G^\bR$ is then an algebra over $D_G^\bR$.
We can define a second $D_G^\bR$-algebra, $F_G^\bR$, which is the linearization of the $D_G$-algebra $F_G$ consisting of those complete derivation trees whose leaves are all labeled by terminals, and where the $D_G$-action is defined by grafting.
There is a forgetful map, $\varphi\colon F_G^\bR \to L_G^\bR$, which sends a derivation tree to its yield.
We now define the ambiguity algebra to be the kernel $\ker\varphi$, i.e.\ those elements of $F_G^\bR$ that are sent to zero by $\varphi$.
\null\hfill$\triangle$
\end{remark}

\subsection{Operads for question decomposition}

We will now apply the language of operads to the context of question decomposition.
Our proposal takes particular inspiration from variants of chain-of-thought prompting, particularly the decomposed prompting approach of \citet{khot2022decomposed} and uncertainty of thoughts approach of \citet{hu2024uncertainty} that decompose input problems to sets of sub-questions.
We first define an operad, $\cQ$, in terms of questions and how they can be de- and re-composed.
Then we will explain how QA models can be interpreted as algebras over $\cQ$.

\begin{definition}
\label{def:Q_sketch}
The \emph{questions operad $\cQ$} consists of the following data:
\begin{itemize}
\item
For every $k \geq 0$, $\cQ(k)$ is the set of questions with $k$ blanks.

\item
For every $k, \ell \geq 0$ and $1 \leq i \leq k$, we define a composition operation
\begin{align}
\circ_i\colon \cQ(k)\times \cQ(\ell) \to \cQ(k+\ell-1)
\end{align}
that plugs the output of the second question into the $i$-th blank of the first.
\null\hfill$\triangle$
\end{itemize}
\end{definition}

It is natural to construct $\cQ$ as a colored operad.
This involves a choice of answer types, defined by the user.
For instance, we could set
\begin{align*}
\mathrm{Color}(\cQ)
\coloneqq
\{\text{time},\text{duration},\text{place},\text{person},\text{other}\}.
\end{align*}

\begin{example}
\label{ex:WW2}
Consider the following questions:
\begin{gather*}
\mathtt{Q1} \coloneqq \text{``When did World War 2 end?''} \in \cQ(;\text{time}), \\
\mathtt{Q2} \coloneqq \text{``Who was President at } -\text{?''} \in \cQ(\text{time};\text{person}), \\
\mathtt{Q3} \coloneqq \text{``Who was } -\text{'s wife?''} \in \cQ(\text{person};\text{person}).
\end{gather*}
The composition $\mathtt{Q3}\circ_1\mathtt{Q2}\circ_1\mathtt{Q1}$ is ``Who was First Lady when World War 2 ended?'' (or a semantic equivalent).
In this case, the associativity of $\cQ$ says that this composition is independent of whether we first compose $\mathtt{Q3}$ and $\mathtt{Q2}$ or $\mathtt{Q2}$ and $\mathtt{Q1}$.
\null\hfill$\triangle$
\end{example}

\begin{remark}
Definition~\ref{def:Q_sketch} should be understood at an abstract level: the composition $\circ_i$ is formal substitution of one typed question template into the $i$-th blank of another.
Operationally, however, one may wish to render such substitutions as fluent natural-language questions, and that surface-realization step will depend on a model.
Accordingly, associativity is best understood up to semantic equivalence.
\null\hfill$\triangle$
\end{remark}

We can interpret question decomposition in terms of composition in $\cQ$.
For instance, consider the question $\mathtt{Q}$, ``How long was it between when the Titanic hit the iceberg and when it sank completely?''
We can decompose $\mathtt{Q}$ into the following tree of questions:

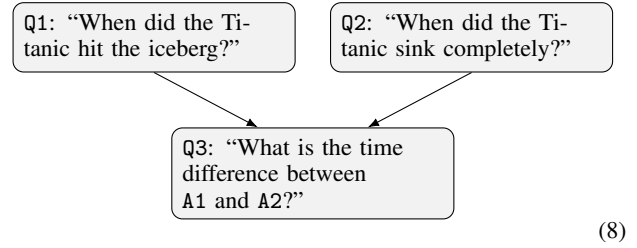
\begin{equation}
\label{eq:titanic_toq}
\footnotesize
\begin{tikzpicture}[
  >={Latex[length=1.5mm]},
  bubble/.style={
    draw, rounded corners=4pt, align=left,
    fill=black!5, inner sep=4pt, text width=3.45cm
  },
  baseline={(current bounding box.center)}
]

\node[bubble] at (0,0) (q3) {%
$\mathtt{Q3}$: ``What is the time difference between\\
  $\mathtt{A1}$ and $\mathtt{A2}$?''
};

\node[bubble] at (-2.1,1.8) (q1) {%
  $\mathtt{Q1}$: ``When did the Titanic hit the iceberg?''
};

\node[bubble] at (2.1,1.8) (q2) {%
  $\mathtt{Q2}$: ``When did the Titanic sink completely?''
};

\draw[->] (q1.south) -- (q3.140);
\draw[->] (q2.south) -- (q3.40);

\end{tikzpicture}
\end{equation}

We can reinterpret this in terms of the questions operad:
\begin{itemize}
\item $\mathtt{Q}$ is an element of $\cQ(;\text{duration})$.
$\mathtt{Q1}$ and $\mathtt{Q2}$ lie in $\cQ(;\text{time})$, and $\mathtt{Q3}$ lies in $\cQ(\text{time},\text{time};\text{duration})$.

\item The fact that \eqref{eq:titanic_toq} is a legitimate decomposition of $\mathtt{Q}$ is witnessed by the fact that $\mathtt{Q1}$, $\mathtt{Q2}$, and $\mathtt{Q3}$ compose to $\mathtt{Q}$ --- that is, $\mathtt{Q} = \mathtt{Q3} \circ_1 \mathtt{Q1} \circ_1 \mathtt{Q2}$.
\end{itemize}

\subsubsection{ML models as algebras over operads}

Now that we have interpreted question decomposition in terms of the questions operad, we can interpret a QA model as an algebra over $\cQ$.
In the following definition, we use the term \emph{value} to mean something that can be used to fill in a blank, or something that can occur as the output of a question.

\begin{definition}
\label{def:V_m}
Suppose that $m$ is a general-purpose question answering model, that takes a question $q$ and returns an answer $v$.
We can use $m$ to define a $\cQ$-algebra, $\cV_m$:
\begin{itemize}
\item As a set, $\cV_m$ consists of all possible values.

\item Given a question $q \in \cQ(k)$ with $k$ blanks, along with $k$ values $v_1,\ldots,v_k$, we can produce a value $v'$ by the following procedure:
\begin{itemize}
\item Fill in the $k$ blanks in $q$ with $v_1,\ldots,v_k$, then process in order to form a question $q'$.

\item Define $v'$ to be the output of $m$ on $q'$.
\null\hfill$\triangle$
\end{itemize}
\end{itemize}
\end{definition}

\subsection{Operadic consistency}

Operads do not only provide a new language for describing question decomposition --- they also point the way toward new methods.
In this subsection, we will define \emph{operadic consistency}, given a model $m$ and a tree of questions $T$.

\subsubsection{Trees of questions and partial collapses thereof}

Fix an instantiation of the questions operad $\cQ$.
By a ``tree of questions'' (or ``ToQ'') $T$, we mean the following:
\begin{itemize}
\item A rooted tree $T$.
By convention, (1) we orient all trees toward the root, (2) each leaf has a single incoming edge, and (3) the root has a single outgoing edge.

\item For every edge $e$ in $T$, a color $t_e \in \mathrm{Color}(\cQ)$.

\item For every vertex $v \in T$, a question $q \in \cQ(t_{e_1},\ldots,t_{e_k};t_{e'})$, where $e_1,\ldots,e_k$ are the incoming edges of $v$ and $e'$ is the outgoing edge of $v$.
\end{itemize}

One example is the ToQ in \eqref{eq:titanic_toq}.
Another is formed by the questions in Ex.~\ref{ex:WW2}, which we use as a running example in this subsection:
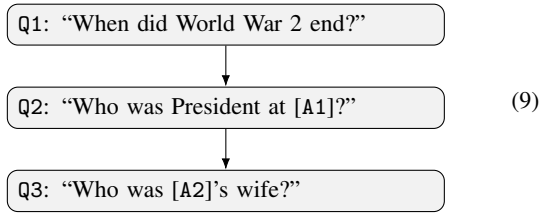
\begin{equation}
\label{eq:ww2_toq}
\footnotesize
\begin{tikzpicture}[
  >={Latex[length=1.5mm]},
  bubble/.style={
    draw, rounded corners=4pt, align=left,
    fill=black!5, inner sep=4pt, text width=5.5cm
  },
  baseline={(current bounding box.center)}
]

\node[bubble] at (0,0) (q3) {%
  $\mathtt{Q3}$: ``Who was [$\mathtt{A2}$]'s wife?''
};

\node[bubble] at (0,1.1) (q2) {%
  $\mathtt{Q2}$: ``Who was President at [$\mathtt{A1}$]?''
};

\node[bubble] at (0,2.2) (q1) {%
  $\mathtt{Q1}$: ``When did World War 2 end?''
};

\draw[->] (q1.south) -- (q2.north);
\draw[->] (q2.south) -- (q3.north);
\end{tikzpicture}
\end{equation}

Any ToQ $T$ has the property that the questions can be composed to form a question with no blanks.
We call this the \emph{total collapse} of $T$.
On the other hand, we can form a \emph{partial collapse} by composing some, but not all, of the questions involved.
Indeed, given a ToQ $T$ and a choice, for every edge, of whether to compose along that edge, we can produce a partial collapse of $T$.
The total number of partial collapses of $T$ is $2^{\#T-1}$, where $\#T$ is the number of vertices of $T$.

\begin{example}
There are 4 partial collapses of \eqref{eq:ww2_toq}:
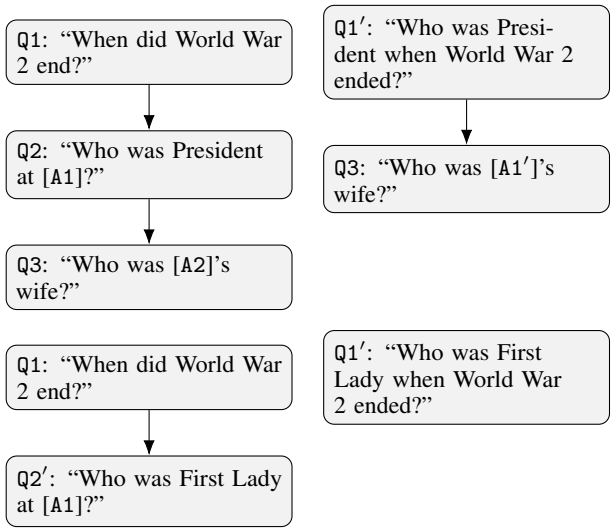
\begin{equation*}
\footnotesize
\begin{tikzpicture}[
  >={Latex[length=2mm]},
  bubble/.style={
    draw, rounded corners=4pt, align=left,
    fill=black!5, inner sep=4pt, text width=3.5cm
  },
  node distance=6mm,
  baseline={(current bounding box.center)}
]

\begin{scope}[shift={(0,0)}]
  \node[bubble] (q1a) {$\mathtt{Q1}$: ``When did World War 2 end?''};

  \node[bubble, below=of q1a] (q2a) {$\mathtt{Q2}$: ``Who was President at [$\mathtt{A1}$]?''};

  \node[bubble, below=of q2a] (q3a) {$\mathtt{Q3}$: ``Who was [$\mathtt{A2}$]'s wife?''};
  \draw[->] (q1a) -- (q2a);
  \draw[->] (q2a) -- (q3a);
\end{scope}

\begin{scope}[shift={(4.2,0)}]
  \node[bubble] (q1'b) {$\mathtt{Q1'}$: ``Who was President when World War 2 ended?''};

  \node[bubble, below=of q1'b] (q3b) {$\mathtt{Q3}$: ``Who was [$\mathtt{A1'}$]'s wife?''};
  \draw[->] (q1'b) -- (q3b);
\end{scope}
\end{tikzpicture}
\end{equation*}
\begin{equation*}
\footnotesize
\begin{tikzpicture}[
  >={Latex[length=2mm]},
  bubble/.style={
    draw, rounded corners=4pt, align=left,
    fill=black!5, inner sep=4pt, text width=3.5cm
  },
  node distance=6mm,
  baseline={(current bounding box.center)}
]

\begin{scope}[shift={(0,0)}]
  \node[bubble] (q1c) {$\mathtt{Q1}$: ``When did World War 2 end?''};

  \node[bubble, below=of q1c] (q2'c) {$\mathtt{Q2'}$: ``Who was First Lady at [$\mathtt{A1}$]?''};
  \draw[->] (q1c) -- (q2'c);
\end{scope}

\begin{scope}[shift={(4.2,0)}]
  \node[bubble] (q1a) {$\mathtt{Q1'}$: ``Who was First Lady when World War 2 ended?''};
\end{scope}
\end{tikzpicture}
\end{equation*}
\null\hfill$\triangle$
\end{example}

\subsubsection{Definition}
\label{sss:operadic_consistency}

Suppose that $T$ is a tree of questions and $m$ is a QA model.
We can use $m$ to answer the questions in $T$: we start at the leaves and work our way toward the root, finally finishing by producing an answer to the root question.
If $\cV_m$ is associative on the nose, then the final answer produced by $m$ on $T$ agrees with $m$'s final answer on the total collapse of $T$.
Moreover, if $T'_1$ and $T'_2$ are two partial collapses of $T$, then the final answers produced by $m$ on $T_1'$ and $T_2'$ must agree.
This leads to our definition of \emph{operadic consistency}.

\begin{definition}
We say that \emph{$m$ is operadically consistent on $T$} if, for any partial collapses $T'_1, T'_2$ of $T$, the final answers produced by $m$ when executed on $T'_1$ and $T_2'$ agree.
\null\hfill$\triangle$
\end{definition}

\begin{example}
Consider $m\coloneqq$ \textsc{Llama 3 8B Instruct}.
When we execute $m$ on the four partial collapses of the ToQ $T$ in \eqref{eq:ww2_toq}, three yield ``Bess Truman'' and one yields ``Eleanor Roosevelt'' --- the total collapse, $m$'s answer to the single-pass question ``Who was First Lady when World War 2 ended?''.
This disagreement witnesses operadic inconsistency.
$\triangle$
\end{example}

\section{Further directions}
\label{s:further_directions}

The framework introduced in this paper opens several directions for future work, both theoretical and experimental.

\paragraph{Extracting operadic structure from chain-of-thought.}
A natural extension is to apply the framework to thinking models, whose extended reasoning traces make the implicit decomposition structure explicit.
This requires first formalizing the model's reasoning as a structured object --- a tree of questions, a more general program, or some other algebraic representation of the trace --- on which a suitable instantiation of operadic consistency can be evaluated.
The depth-2 chain extraction used in our companion paper \citep{bottman:operadic_consistency_empirical} is a minimal first attempt; richer formalisms open the possibility of consistency checks that capture more of the trace's compositional content.

\paragraph{Cohomological invariants.}
Algebras over operads admit a notion of \emph{cohomology}, and we believe that the cohomology of the $\cQ$-algebra associated to a QA model may carry meaningful information about the structure of a model's inconsistencies --- distinguishing, for instance, between correctable inconsistencies and more fundamental failures of multi-step reasoning.

\paragraph{Other algebras over $\cQ$.}
The operadic perspective suggests that question decompositions need not be interpreted only by their final answers.
They may also be interpreted by likelihoods, confidence scores, evidence bundles, costs, latent representations, or robustness profiles, each giving an algebra over the same underlying compositional structure.

\paragraph{Companion paper.}
A first empirical installment of this program already exists: our companion paper \citep{bottman:operadic_consistency_empirical} evaluates operadic consistency across twelve instruction-tuned LLMs and four multi-hop QA datasets, where it is strongly correlated with accuracy and yields selective-prediction improvements over temperature-based self-consistency at equal inference cost. We refer the reader there for protocols, baselines, and full results.

\section*{Acknowledgements}

N.B.\ was supported by the Defense Advanced Research Projects Agency (DARPA) through the Artificial Intelligence Quantified (AIQ) program, under Cooperative Agreement HR00112520028.
The views, opinions, and/or findings expressed are those of the authors and should not be interpreted as representing the official views or policies of the Department of Defense or the U.S.\ Government.

N.B.\ thanks AI2 for its hospitality during this collaboration.

\section*{Impact Statement}

This paper develops a mathematical framework for question decomposition in language models.
We foresee no direct negative societal impacts; operadic consistency could in principle be used to audit deployed QA systems.

\printbibliography

\end{document}